%% file: acl.tex
\pdfoutput=1

\documentclass[11pt]{article}

\usepackage[]{acl}

\usepackage{times}
\usepackage{latexsym}

\usepackage[utf8]{inputenc}

\usepackage{microtype}

\input{math_commands.tex}

\usepackage{url}
\usepackage{cleveref}
\usepackage{xcolor}
\usepackage[linesnumbered,ruled,vlined]{algorithm2e}
\usepackage{graphicx}
\usepackage{float}
\usepackage{adjustbox}
\usepackage{tabularx}
\usepackage{multirow}
\usepackage{graphicx}
\usepackage{subcaption}
\newcommand{\xhdr}[1]{\noindent{\bfseries #1}   }
\graphicspath{ {./images/} }

\SetKwInput{KwInput}{Init}      
\SetKwInput{KwOutput}{Output}   

\DeclareMathOperator*{\ccat}{||}
\DeclareMathOperator*{\sstack}{\texttt{stack}}

\title{LAGr: Label Aligned Graphs for Better Systematic Generalization in Semantic Parsing}

\author{Dóra Jámbor\thanks{Corresponding author. Work partly done during an internship at ServiceNow Research.} \\
Mila - Quebec AI Institute\\
McGill University\\
\texttt{dora.jambor@mail.mcgill.ca}
\And 
Dzmitry Bahdanau \\
ServiceNow Research\\
Mila - Quebec AI Institute\\
McGill University\\
Canada CIFAR AI Chair \\
}

\begin{document}
\maketitle
\begin{abstract}
    Semantic parsing is the task of producing structured meaning representations for natural language sentences. 
    Recent research has pointed out that the commonly-used sequence-to-sequence (seq2seq) semantic parsers struggle to generalize systematically, i.e. to handle examples that require recombining known knowledge in novel settings. In this work, we show that better systematic generalization can be achieved by producing the meaning representation directly as a graph and not as a sequence. To this end we propose LAGr (\textbf{L}abel \textbf{A}ligned \textbf{Gr}aphs), a general framework to produce semantic parses by independently predicting node and edge labels for a complete multi-layer input-aligned graph. The strongly-supervised LAGr algorithm requires aligned graphs as inputs, whereas weakly-supervised LAGr infers alignments for originally unaligned target graphs using approximate maximum-a-posteriori inference. 
    Experiments demonstrate that LAGr achieves significant improvements in systematic generalization upon the baseline seq2seq parsers in both strongly- and weakly-supervised settings.
\end{abstract}

\section{Introduction}
Recent research has shown that neural models struggle to systematically generalize to examples with unseen combinations of seen rules from the training set \citep{lake_generalization_2018,finegan-dollak_improving_2018,hupkes_compositionality_2019}.
Systematic generalization is especially important for the task of semantic parsing, which requires models to translate natural language sentences to structured meaning representations (MRs), such as SPARQL database queries or lambda calculus logical forms.
To generalize systematically in this task, the model must be capable of producing MRs for examples that feature new combinations of meaning construction rules, such as the rule that maps a noun like ``\textit{hedgehog}'' in Figure \ref{fig:cogs_2examples} to its respective predicate $hedgehog(.)$, and the rule that defines which semantic role with respect to the verb  (e.g. \textit{agent} or \textit{theme}) the resulting predicate takes.
Using synthetic \citep{bahdanau_closure_2019,kim_cogs_2020,keysers_measuring_2020}  and natural benchmarks \citep{finegan-dollak_improving_2018,shaw_compositional_2020}, researchers have been studying systematic generalization of existing semantic parsing methods as well as proposing new approaches such as using meta-learning \citep{conklin_meta-learning_2021}, pretrained models \citep{furrer_compositional_2020}, or intermediate meaning representations \citep{herzig_unlocking_2021}.

\begin{figure}
    \begin{tabular}{l}
    \textbf{Training example} \\ \hline
    A \textcolor{red}{\textit{\textbf{hedgehog}}} ate the cake \\
    $*{hedgehog}(x_1) \wedge cake(x_4) \wedge $\\ $eat.agent(x_2, {x_1}) \wedge eat.theme(x_2, x_4)$\\ \hline
    \textbf{Generalization example} \\ \hline
    The baby liked the \textcolor{red}{\textit{\textbf{hedgehog}}}\\
    $*baby(x_1) \wedge {hedgehog}(x_4) \wedge $\\ $like.agent(x_2, x_1) \wedge like.theme({x_2, x_4})$) \\ \hline
\end{tabular}
    \caption{Examples from the training and the generalization sets of the COGS dataset \citep{kim2020cogs}. While ``\textit{hedgehog}'' is only observed in the \textit{agent} role during training, the generalization set features this word in the \textit{theme} role.}
    \label{fig:cogs_2examples}
\end{figure}

The dominant framework in these studies is \textit{sequence-to-sequence} \citep[seq2seq, ][]{sutskever_sequence_2014,bahdanau_neural_2015} learning, whereby the model produces a serialized MR in an autoregressive fashion, by predicting one token at a time, while conditioning on all previously generated tokens. We hypothesize that for semantic parsing constructing the MR by combining independent predictions that are not conditioned on each other can generalize more systematically than seq2seq. For example, consider the sentence ``\textit{The dog liked that the hippo danced}''. Arguably, the predictions that ``\textit{dog}'' is the agent of ``\textit{like}'' and that ``\textit{hippo}'' is the agent of ``\textit{danced}'' can be made independently of each other. Our intuition is that a model that predicts such aspects of meaning independently of each other can be better at learning context-insensitive rules because the overall context for each individual prediction is reduced. 

\begin{figure*}
  \begin{subfigure}{0.45\textwidth}
    \includegraphics[width=\linewidth]{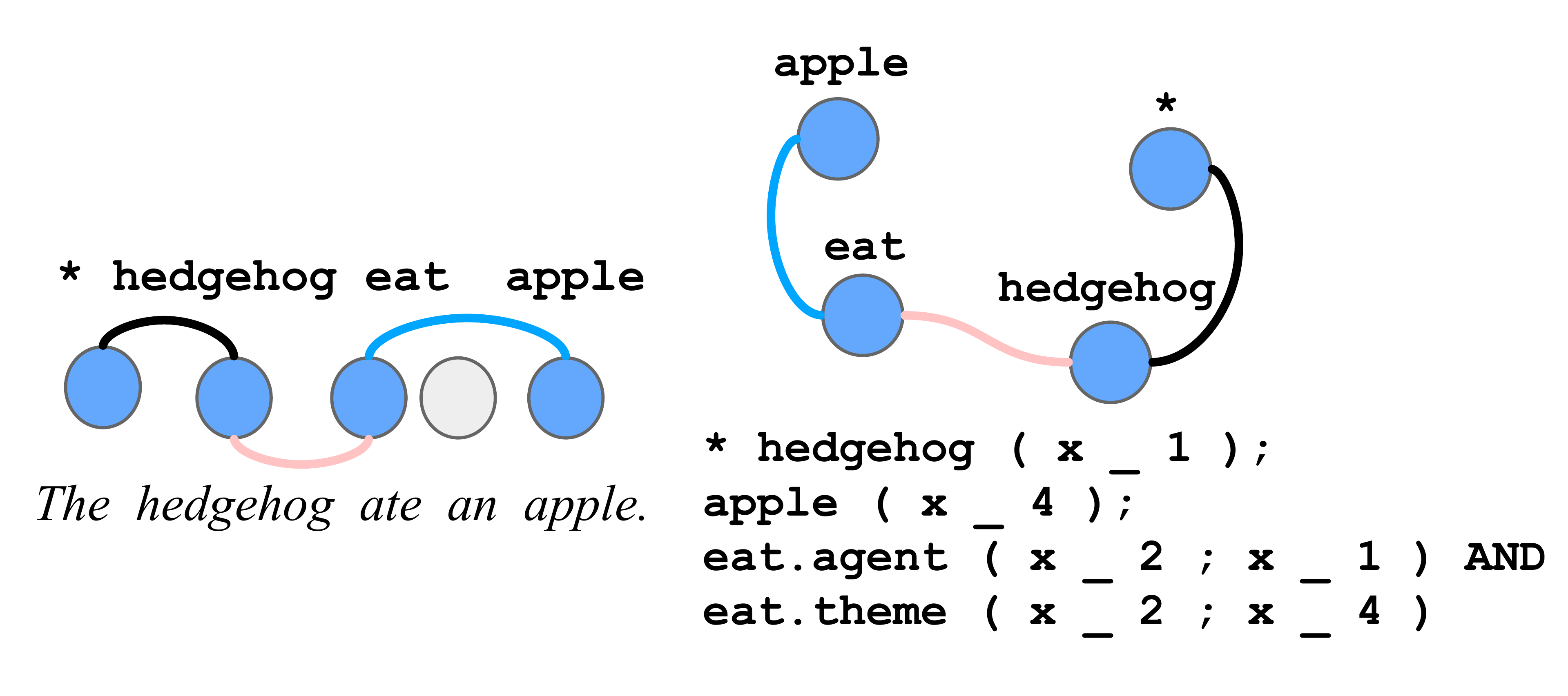}
    \caption{COGS} \label{aligned_graph:1a}
  \end{subfigure}%
  \hspace*{\fill}   
  \begin{subfigure}{0.55\textwidth}
    \includegraphics[width=\linewidth]{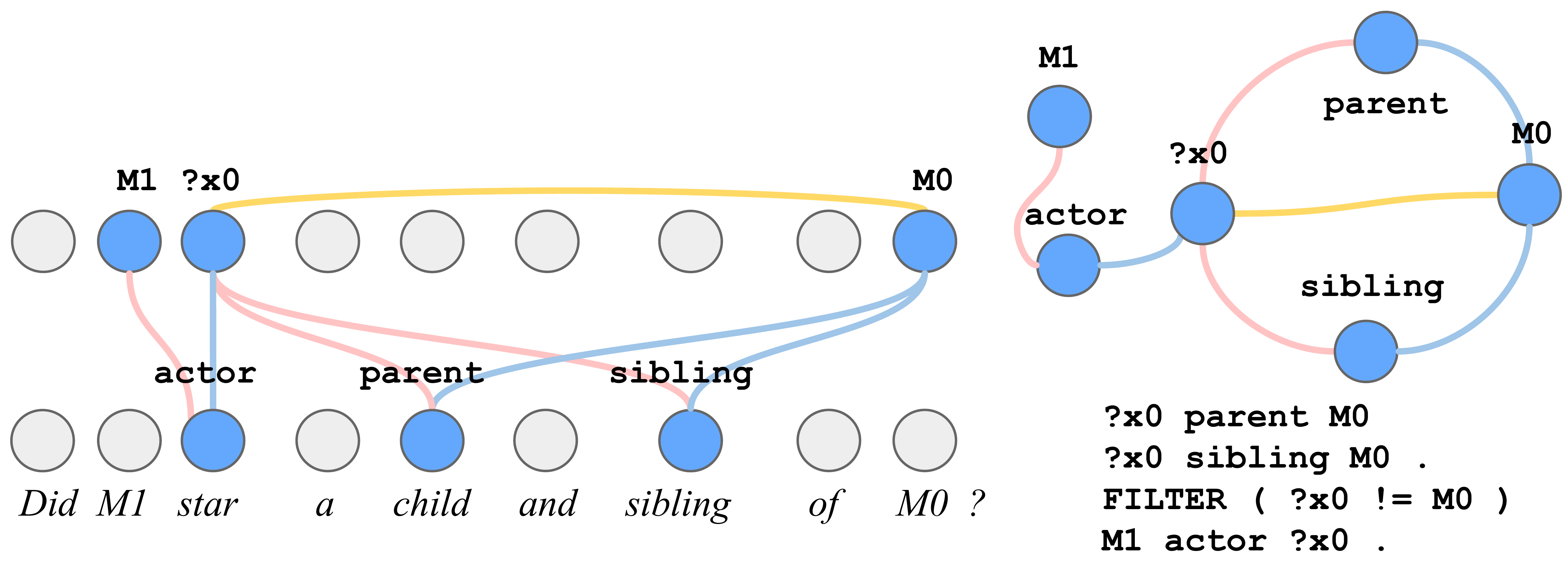}
    \caption{CFQ} \label{aligned_graph:1b}
  \end{subfigure}
  \caption{Aligned and unaligned graphs for COGS (a) and CFQ (b). For COGS, pink, blue and black denote \color{pink}\textbf{\texttt{agent}}\color{black}, \color{blue}\texttt{theme} \color{black} and \textbf{\texttt{article}} \color{black}  edges, respectively. For CFQ, yellow, pink and blue mark \color{yellow} \textbf{\texttt{FILTER}}\color{black}, \color{pink}\textbf{\texttt{agent}}\color{black}, \color{blue}\texttt{theme} \color{black} edges. Grey nodes mark \texttt{null} nodes, and \texttt{*} denotes the definite article. The aligned graph for CFQ is provided for illustration purposes, and was not used for training. See Section \ref{sec:experiments} for the learned aligned graphs.}
\label{aligned_graphs}
\end{figure*}

Following this intuition,  we propose LAGr (\textbf{L}abel \textbf{A}ligned \textbf{Gr}aphs), a framework to produce semantic parses by independently labelling the nodes and edges of a fully-connected multi-layer output graph that is aligned with the input utterance.
While the general idea of predicting semantic parses as graphs is not new \citep{lyu_amr_2018}, the systematic generalization benefits of doing so have not been investigated prior to this work.
Importantly, LAGr retains most of the flexibility that seq2seq models have, without the complexity and rigidity that comes with other alternatives to seq2seq, such as grammar-based methods \citep{herzig_span-based_2020}.

We first introduce LAGr in the strongly-supervised setting where output graphs are aligned to the input sequences, thus allowing for standard supervised training. For the weakly-supervised case when the alignment is not available, we treat it as a latent variable. We infer the latent alingment with a simple and novel approximate maximum-a-posteriori (MAP) inference approach which involves solving several minimum cost bipartite matching problems with the Hungarian algorithm \citep{kuhn_hungarian_1955}. We then use the resulting aligned graphs to train the model. 
Our experiments demonstrate that in both strongly- and weakly-supervised settings LAGr significantly improves upon comparable seq2seq semantic parsers on the COGS and CFQ datasets \citep{kim_cogs_2020,keysers_measuring_2020}. 

\section{Semantic Parsing by Labeling Aligned Graphs}
\label{lagr_methodology}
We present LAGr (\textbf{L}abel \textbf{A}ligned \textbf{G}raphs), a framework for constructing meaning representations (MR) directly as graphs (i.e., \textit{MR graphs}).
When LAGr is used to output logical forms, the graph nodes can be variables, entities, categories and predicates, and graph edges can be the Neo-Davidsonian style semantic role relations that the nodes appear in, e.g. \textit{``is-agent-of''} or \textit{``is-theme-of''} \citep{parsons1990events}. 
While this work focuses on predicting logical forms, LAGr can, in principle, also be used to output other kinds of graphs, such as abstract syntax tree parses of SQL queries.
As illustrated in Figure \ref{aligned_graphs}, LAGr predicts the output by labeling the nodes and edges of a fully-connected multi-layer output graph that is aligned with the input utterance. 
We label a multi-layer as opposed to a single-layer graph because some MR graphs have more nodes than the number of input tokens (see Section \ref{cfq_experiments} for an example).

\paragraph{Notation and Terminology}
Formally, let $x = x_1, x_2, ..., x_N$ denote a natural language utterance of $N$ tokens. 
LAGr produces an MR graph $G$ by labeling the nodes and edges of a complete graph $\Gamma_\mathrm{a}$ with $M = L \cdot N$ nodes that are arranged in $L$ layers. The layers are aligned with the input sequence $x$ in a way that for each input position $i$ there is a unique corresponding output node in each layer. We say that nodes from different layers that are aligned with the position $i$ form a column (an example column in Figure \ref{aligned_graph:1b} contains the nodes labeled as \texttt{actor} and \texttt{?x0} for the word \textit{star} at the position $i=3$).

We write $\Gamma_\mathrm{a} = (z, \xi)$ to indicate that a complete labeled graph $\Gamma_\mathrm{a}$ is characterized by its node labels $z \in V_n^M$ and edge labels $\xi \in V_e^{M \times M}$, where $V_n$ and $V_e$ are node and edge label vocabularies, respectively.
Both vocabularies also include additional \texttt{null} labels that we use as padding (e.g. grey nodes in Figure \ref{aligned_graphs} are labeled as \texttt{null}). 
To produce the output MR graph $G$ from $\Gamma_\mathrm{a}$, we remove all \texttt{null} nodes and \texttt{null} edges.
Lastly, we use $z_j$ and $\xi_{jk}$ notations to refer to the labels of node $j$ and of the edge $(j, k)$ where $j = (l - 1)N + i$ is a one-dimensional index that corresponds to the $i$-th node in the $l$-th layer. 

\subsection{Labeling Aligned Graphs}
\label{stronglysup_lagr}

To label the nodes of $\Gamma_\mathrm{a}$ we encode the input utterance $x$ as a matrix of $N$ $d$-dimensional vectors $H=f_{enc}(x) \in \mathbb{R}^{N \times d}$, 
where $f_{enc}$ can be an arbitrary encoder model such as LSTM \citep{hochreiter1997long} or a Transformer \citep{vaswani_attention_2017}. 
LAGr then defines a factorized distribution $p(z|x)$ over the node labels~$z$ as follows:
\begin{align}
    O = \ccat\limits_{l=1}^{L} H W^l, \\
    \mathcal{\pi} = \texttt{softmax}(O), \\
    p(z|x) = \prod\limits_{j=1}^M p(z_j|x) = \pi_{j,z_j}, \label{node_predications}
\end{align}
where $O \in \mathbb{R}^{M \times |V_n|}$ contains logits for $M=N\times L$ nodes from all the $L$ graph layers, $\ccat$ denotes the concatenation operation along the node axis, $W^l$ denotes the weight matrix for layer~$l$.
Here and in following equations $\texttt{softmax}(.)$ is applied to the last dimension of the input tensor and every multiplication by a weight matrix is followed by the addition of a bias vector which we omit to enhance clarity. 
Our edge labelling computation is reminiscent of the multi-head self-attention by \citet{vaswani_attention_2017}, with the key difference that softmax is applied across the edge labels and not across positions:
\begin{gather*}
    H^{\alpha}_q = \ccat\limits_{l=1}^L H U^{\alpha,l}, \phantom{==} H^{\alpha}_k = \ccat\limits_{l=1}^L H V^{\alpha,l}, \\
    \rho  = \texttt{softmax} \left[ 
        \sstack\limits_{\alpha \in V_e}  \left[
            H^{\alpha}_q {H^{\alpha}_k}^T
        \right]
    \right],
\end{gather*}
where $H^{\alpha}_q$ and $H^{\alpha}_k$ contain concatenated key and query vectors for the label $\alpha \in V_e$ across all $L$ graph layers,
$U^{\alpha,l}, V^{\alpha,l} \in \mathbb{R}^{\frac{d}{|V_e|}, \frac{d}{|V_e|}}$ are the weights for the edge label $\alpha$, and the $\sstack$ operator stacks the matrices into a 3D tensor to which $\softmax$ is subsequently applied.
Similarly to $p(z|x)$, we obtain $p(\xi|x)$ as follows:
\begin{equation}
p(\xi|x) = \prod\limits_{j=1}^M \prod\limits_{k=1}^M p(\xi_{jk}|x)
    = \prod\limits_{j=1}^M \prod\limits_{k=1}^M
    \rho_{jk\xi_{jk}}. \label{edge_predications}
\end{equation}
The factorized nature of Equations \ref{node_predications} and \ref{edge_predications} makes the argmax inference $\hat{z}, \hat{\xi} = \argmax p(z, \xi|x)$ trivial to perform. 
When the groundtruth aligned graph $\Gamma_\mathrm{a}^{\ast}=(z^*, \xi^*)$ for the MR graph $G$ is available, LAGr can be trained by directly optimizing $\log p(z=z^{\ast}, \xi=\xi^{\ast}|x)$. We refer to this training setting as \textit{strongly-supervised LAGr}. 

\subsection{Weakly-supervised LAGr}
In many practical settings, the alignment between the MR graph $G$ and the sequence $x$ is unavailable, making the aligned graph $\Gamma_\mathrm{a}$ unknown.
To address this common scenario, we propose a \textit{weakly-supervised LAGr} algorithm based on a latent alignment model. Similarly to the strongly-supervised case, we assume that the MR graph can be represented as a labeled complete, multi-layer graph $\Gamma_\mathrm{na}=(s \in V_n^M, e \in V_e^{M \times M})$, with the difference that in this case the alignment between $x$ and $\Gamma_\mathrm{na}$ is not known. 
We assume a generative process whereby $\Gamma_\mathrm{na}$ is obtained by permuting the columns of the latent aligned graph $\Gamma_\mathrm{a}$ with a random permutation $a$, where $a_j$ is the index of the column in $\Gamma_\mathrm{a}$ that becomes the $j$-th column in $\Gamma_\mathrm{na}$. For the rest of this section we focus on the single layer ($L=1$) case to simplify the formulas. For this case our probabilistic model defines the following distribution over $\Gamma_\mathrm{na}=(s, e)$:
\begin{align}
\begin{split}
    p(e, s\vert x) &= \sum_a \sum_z \sum_\xi p(e, s, a, z, \xi | x) \\
    &= \sum_a p(a) \prod_j p(z_{a_j}=s_j|x) \\
    &\prod_j \prod_k p(\xi_{a_ja_k}=e_{jk}|x), \label{inf}
\end{split}
\end{align} 
where $p(a) = 1/N!$. 
Computing $p(e, s|x)$ exactly is intractable. For this reason,  we train LAGr by using an approximation of $p(e, s|x)$ in which instead of summing over all possible aligments $a$, we only consider the maximum-a-posteriori (MAP) alignment $\hat{a} = \argmax_a p(a|e, s, x)$. This approach is sometimes called the hard Expectation-Maximization algorithm in the literature on probabilistic models \citep{svensen2007pattern}. The training objective thus becomes
\begin{gather*}
\begin{split}
    p(e, s\vert \hat{a}, x) &= \\
    & \hspace{-1.5cm} \prod_j p(z_{\hat{a}_j}=s_j|x) \prod_j \prod_k p(\xi_{\hat{a}_j,\hat{a}_k}=e_{jk}|x).
\end{split}
\end{gather*}
To infer the MAP alignment $
\hat{a}$, we need to solve the following inference problem:
\begin{gather}
\begin{split}
\hat{a} &= \argmax_a p(a|e,s,x) \\ 
&= \argmax_a \log p(s|a, x) + \log p(e|a, x) \label{eq:posterior}\\
&= \argmax_a \Big[\sum_j \log p(z_{a_j}=s_j|x) \\
&\phantom{=}\, + \sum_j \sum_k \log p(\xi_{a_j,a_k}=e_{j,k}|x) \Big]
\end{split}
\end{gather}
We are not aware of an exact algorithm for solving the above optimization problem, however if the edge log-likelihood term $\log p(e|a,x)$ is dropped in the equations above, maximizing the node label probability $p(s|a,x)$ is equivalent to a standard
minimum cost bipartite matching problem. This optimization problem can be solved by a polynomial-time Hungarian algorithm \citep{kuhn1955hungarian}. We can thus use an approximate MAP alignment $\hat{a}^1=\argmax_a \sum_j \log p(z_{a_j}=s_j|x)$. While dropping $p(e|a, x)$ from Equation \ref{eq:posterior} is a drastic simplification, in situations where node labels $s$ are unique and the model is sufficiently trained to output sharp probabilities $p(z_j|x)$ we expect $\hat{a}^1$ to often match $\hat{a}$. To further improve the MAP alignment approximation and alleviate the reliance on the node label uniqueness, we generate a shortlist of $K$ candidate alignments by solving $K$ noisy matching problems of the form $\argmax_a \sum_j \log p(z_{a_j}=s_j|x) + \epsilon_{ja_j}$, where $\epsilon_{ja_j}\sim N(0, \sigma)$. We then select the alignment candidate $a$ that yields the highest full log-likelihood $\log p(s|a, x) + \log p(e|a, x)$.

We refer the reader to Algorithm \ref{lagr} for a detailed presentation of weakly-supervised LAGr. 

\begin{algorithm}
\DontPrintSemicolon
  
\KwInput{Let $K$ be the number of alignment candidates, $T$ be the number of training steps, and ${\theta}_t$ be the model parameters after $t$ steps.}
\For{t=1, ..., T}    
{ 
	sample example $(x, e, s)$ \\ 
	\For{$\kappa$=1, ..., K}    
    {
        $\epsilon_{ji} \sim N(0, \sigma)$ \\
        $cost_{ji} = - \log p(z_i=s_j|x) + \epsilon_{ji}$\\
        $a^{\kappa} = \texttt{MinCostMatch} (cost)$ \\
        $J^{\kappa} = \sum_j \log p(z_{a^{\kappa}_j}=s_j|x)$ \\
        $\phantom{===} +\sum_j \sum_k \log p(\xi_{a^{\kappa}_ja^{\kappa}_k}=e_{jk}|x)$ \\
    }
    $\hat{\kappa} = \argmax_{\kappa} J^{\kappa}$ \\
	$\theta_{t+1} \leftarrow \texttt{Optimizer}(\theta_t, \nabla_{\theta}-J^{\hat{\kappa}})$
}
\textbf{return} $\theta_{T+1}$
\caption{\label{lagr}Training LAGr with weak supervision}
\end{algorithm}

\section{Related Work}
\label{sec:related_work}
The LAGr approach is heavily inspired by graph-based dependency parsing algorithms \citep{mcdonald_discriminative_2006}. In neural graph-based dependency parsers \citep{kiperwasser_simple_2016,dozat_deep_2017} the model is trained to predict the existence and the label of each of the possible edges between the input words. The Abstract Meaning Representation (AMR) parser by \citet{lyu_amr_2018} brings similar methodology to the realm of semantic parsing, although they do not consider the systematic generalization implications of using a graph-based parser instead of a seq2seq one. \citet{lyu_amr_2018} only output single layer graphs which requires aggresive graph compression; in LAGr we allow the model to output a multiple layer graph instead. Lastly, the amortized Gumbel-Sinkhorn alignment inference used by \citet{lyu_amr_2018} is much more complex than the Hungarian-algorithm-based approximate MAP inference that we employ here. Another important inspiration for LAGr is the UDepLambda method \citep{reddy_transforming_2016} that converts dependency parses into graph-like logical forms. LAGr can be seen as an algorithm that produces UDepLambda graphs directly with the neural model, side-stepping the intermediate dependency parsing step.

Another alternative to seq2seq semantic parsers are span-based parsers that predict span-level actions for building MR expressions from sub-expressions \citep{pasupat_span-based_2019,herzig_span-based_2020,liu_learning_2021}. A prerequisite for using a span-based parser is an MR that can be viewed as a recursive composition of MRs for subspans. While this strong compositionality assumption holds for the logical forms used in earlier semantic parsing research (e.g. \citet{zettlemoyer_learning_2005}), an intermediate MR would be required to produce other meaning representations, such as e.g. SPARQL or SQL queries, with a span-based parser. The designer for an intermediate MR for a span-based parser must think about MRs for spans and how they should be composed. This can sometimes lead to non-trivial corner cases, such as e.g. ternary grammar rules in \citet{herzig_span-based_2020}. On the contrary, a graph-based parser can in principle produce any graph, although in practice in our experiments we compress the raw graphs slightly to make the learning problem easier. 

Other related semantic parsing approaches include the semantic labeling method by \citet{zheng2020compositional} and the structured reordering approach by \citet{wang_structured_2021}.
\citet{zheng2020compositional} show that labelling the input sequence prior to feeding it to the seq2seq semantic parser improves systematic generalization. Compared to that study, our work goes one step further by adding edge labeling, which allows us to let go of the seq2seq model entirely. \citet{wang_structured_2021} model semantic parsing as structured permutation of the input sequence followed by monotonic segment-level transduction. This approach achieves impressive results, but is considerably more complex than LAGr. Finally, \citet{guo_hierarchical_2020} achieve a very high performance on CFQ by combining the sketch prediction approach \citep{dong_coarse--fine_2018} with an algorithm that outputs the MR as a directed acyclic graph (DAG). Unlike LAGr, their algorithm produces the DAG in a sequential left-to-right fashion. Notably, the non-hierachical version of this algorithm without sketch prediction performs poorly. 

Concurrently with this work, \citet{ontanon_making_2021} show that semantic parsing by sequence tagging improves systematic generalization. Their sequence tags are similar to the aligned graphs that we predict with LAGr when using a single graph layer. \citet{ontanon_making_2021} do not discuss how to infer sequence tags from logical forms when the former are not available.

\section{Experiments}
\label{sec:experiments}
We demonstrate the effectiveness of LAGr on two systematic generalization benchmarks for semantic parsing: COGS \citep{kim_cogs_2020} and Compositional Freebase Questions (CFQ, \citet{keysers_measuring_2020}) \footnote{Our code is available under \url{https://github.com/ElementAI/lagr}}.

\subsection{COGS}
\begin{table*}
\begin{center}
{\small 
\begin{tabular}{lccc}
& \multicolumn{3}{c}{Exact match accuracy (\%)} \\ \hline
& train & test & gen \\ \hline
LSTM+Attn  $\diamondsuit$ &  - & 99. & 16. ($\pm$8.)    \\
Transformer  $\diamondsuit$ &  - & 96. & 35. ($\pm$6.)    \\
LSTM+Attn $\heartsuit$ &  - & -   & 51. ($\pm$5.)  \\
Transformer $\clubsuit$ &   -  &  - & \textbf{81.} ($\pm$1.)       \\ 
LSTM + Lex: Simple $\heartsuit$                     &   - & - &  \textbf{82. }($\pm$1.)   \\ 
LSTM + Lex: PMI $\heartsuit$                        &   - & - &  \textbf{82. }($\pm$0.)    \\ 
LSTM + Lex: IBMM2 $\heartsuit$                      &   - & - &  \textbf{82. }($\pm$0.)    \\ 
\hline
LSTM+Attn (ours)                                     & 100 ($\pm$0.0)  & 99.6 ($\pm$0.2)  & 26.1 ($\pm$6.8)   \\
LSTM$_{sh}$ strongly-supervised LAGr      & 100 ($\pm$0.0)    & 99.9 ($\pm$0.1)             & 39.0 ($\pm$9.1) \\
LSTM$_{sep}$ strongly-supervised LAGr            & 100 ($\pm$0.0)  & 100 ($\pm$0.0)    & 71.4 ($\pm$2.9)   \\
Transformer (ours)                      & 100 ($\pm$0.0)   & 99.8 ($\pm$0.0)   & 80.6 ($\pm$1.4) \\
Transformer$_{sh}$ strongly-supervised LAGr        &  100 ($\pm$0.0)  & 100 ($\pm$0.0)  &  80.2 ($\pm$1.4)  \\ 
Transformer$_{sep}$ strongly-supervised LAGr &  100 ($\pm$0.0)  & 99.9 ($\pm$0.1) &  \textbf{82.5} ($\pm2.9$)         \\ 
Transformer$_{sep}$ weakly-supervised LAGr &  100 ($\pm$0.0)  & 99.9 ($\pm$0.0) & 80.7 ($\pm$2.5)  \\
Transformer$_{sep}$ weakly-supervised LAGr + Retrain & 100 ($\pm$0.0)  & 99.9 ($\pm$0.0) & \textbf{82.3} ($\pm$2.3) \\ \hline 
\end{tabular}}
\end{center}
\caption{Average exact match accuracy and standard deviation on COGS. \textbf{Bottom}: reproduced seq2seq baselines and LAGr.
\textbf{Middle:} Seq2seq baselines including the original results by \citet{kim_cogs_2020} $\diamondsuit$, best Transformer baseline by \citet{csordas2021devil} $\clubsuit$, and the best LSTM baseline by \citet{akyurek_lexicon_2021} $\heartsuit$. We also show results by the lexicon-based approach by \citet{akyurek_lexicon_2021}.
}
\label{cogs_results}
\end{table*}

\xhdr{Dataset} 
COGS \citep{kim_cogs_2020} is a semantic parsing benchmark that requires models to translate English sentences to Neo-Davidsonian lambda calculus logical forms.
As shown in Figure \ref{fig:cogs_2examples}, the out-of-distribution generalization set of COGS features novel combinations of words and syntactic structures from the training dataset (more examples available in Appendix \ref{cogs_examples}).

\xhdr{Graph Construction}
In order to study LAGr on COGS, we first convert the logical forms to UDepLambda-style \citep{reddy_transforming_2016} MR graphs.
Specifically, we construct the graph nodes using the one- and two-place predicates and definite articles (e.g. \texttt{hedgehog, apple, eat} and the \texttt{*} nodes in Figure \ref{aligned_graph:1a}). 
We do not create dedicated nodes for variables, as every variable in COGS is either an argument to a unique one-place predicate (e.g. $x_1$ is for $\texttt{hedgehog}(x_1)$), or the first argument to a unique two-place predicate (e.g. $x_2$ for \texttt{eat} in $\texttt{eat.agent}(x_2, x_1)$). Instead, we let the respective predicate node represent the variable.
The labeled edges for our graphs are defined by the Neo-Davidsonian role predicates of the logical forms (such as \texttt{agent}, \texttt{theme}, \texttt{recipient}, \texttt{ccomp}, \texttt{nmod.on}, \texttt{nmod.in}, \texttt{xcomp}, \texttt{nmod.beside}). For example, the conjunct $\texttt{eat.agent}(x_2, x_1)$ results in an \texttt{agent} edge between the \texttt{eat} and \texttt{hedgehog} nodes. We also add special \texttt{article} edges to connect definite article nodes (denoted by the \texttt{*} label) to their respective nouns (e.g. \texttt{hedgehog} in Figure \ref{aligned_graph:1a}).
We take advantage of the correspondence between variable names and input positions ($x_i$ corresponds to the $i$-th token) to construct single-layer ($L=1$) aligned graphs $\Gamma_\mathrm{a}$ for COGS that are suitable for strongly-supervised LAGr, as described in Section \ref{stronglysup_lagr}. The node and edge vocabularies for the aligned graphs contain 645 and 10 labels respectively, each including a \texttt{null} label. 

\xhdr{Training Details}
Hyperparameter tuning on COGS is challenging since the performance on the in-distribution development set always saturates to near 100\%.
We adopt the hyperparameter tuning procedure discussed in \citet{conklin_meta-learning_2021} to find the best configuration for our baselines and strongly-supervised LAGr models.
Specifically, we create a ``Gen Dev'' dataset by sampling 1000 random examples from the generalization set and use them to find the best hyperparameter configuration. 
We find that our Transformer-based seq2seq and LAGr models perform better when embeddings are initialized following \citet{he2015delving} and when positional embeddings are scaled down by $\frac{1}{\sqrt{dim}}$. The latter techniques were adopted following the recent work of \citet{csordas2021devil} under the PED (Positional Embedding Downscaling) name. We report the exact match accuracy, i.e., the percentage of examples for which the predicted graphs after serialization yielded the same logical form, as well as the standard deviation over at least 10 random seeds.
We tune the hyperparameters for strongly-supervised LAGr first; 
we then use the same configuration for weakly-supervised LAGr and only tune the inference hyperparameters, i.e. the number of candidates $K$ and the noise level $\sigma$. Since weakly-supervised LAGr does not always converge on the training set,
we implement a restart mechanism that relaunches experiments with a new random seed where a training performance of at least 95\% is not achieved.
Setting $K=10$ and $\sigma=1.0$ allows us to achieve a convergence rate of around 50\%. 
For more details on our hyperparameter search, and best configurations, we refer the reader to Appendix \ref{cogs_gen_dev}.

Additionally, we observe that the training loss does not go to 0 in the weakly-supervised setting. We attribute this to a significant (2.7\%) percentage of training examples in which there are three and more nodes with the same label (namely ``*'' for definite articles), which presents a challenge to our alignment inference mechanism. To remedy this, we cache and append the previously used alignment as the $K+1$st alignment candidate (see lines 3-8 in Algorithm \ref{lagr}). This allows the model to remember low-loss alignments and thereby helps achieve full convergence. 
Lastly, we also run weakly-supervised LAGr with retraining, in which we take the final learned alignments for all examples and retrain models with the learned alignments being used as strong supervision.

\xhdr{Baselines}
We compare LAGr to LSTM- and Transformer- based seq2seq semantic parsers that produce logical forms as sequences of tokens. 
In addition to training our own seq2seq baselines, we also include baseline results from the original COGS paper by \citet{kim_cogs_2020} and from follow-up works by \citet{akyurek_lexicon_2021}, and \citet{csordas2021devil}.
We also compare LAGr to a lexicon-based seq2seq model ``LSTM+Lex'' by \citet{akyurek_lexicon_2021} that leverages the copy mechanism in the seq2seq decoder to perform a lexical lookup to generate the output token.

\xhdr{Results} 
\Cref{cogs_results} shows that our best Transformers trained with LAGr outperform the original (35\% from \citet{kim2020cogs} and 81\% from \citet{csordas2021devil}) and our reproduced (80.6\%) seq2seq Transformer baselines, obtaining 82.5\% and 82.3\% exact match accuracy in the strongly- and weakly-supervised settings, respectively. 

We experiment with two variations of LAGr: using shared encoders and separating encoders for syntax (i.e., node predictions) and semantics (i.e., edge predictions) --- reflected in \Cref{cogs_results} by the subindex "$\_sh$" versus "$\_{sep}$" in the model names respectively.
We achieve the best result in the strongly-supervised setting using separate encoders. 
While this setting significantly improves the performance of LAGr in all cases, for the strongly-supervised LSTM-based LAGr models, separating encoders seems to be crucial (71.4\% vs 39.0\%).

The use of retraining in weakly-supervised LAGr is helpful. It allows us to increase the accuracy of weakly-supervised LAGr to match our strongly-supervised result.
Finally, LAGr is able to match the performance of the LSTM+Lex approach by \citet{akyurek_lexicon_2021} without relying on the use of lexicons --- a result we further discuss in Section \ref{sec:discussion}.

\subsection{CFQ}
\begin{table*}
\centering
\begin{adjustbox}{width=\textwidth}
\begin{tabular}{lcccccc}
     & \multicolumn{6}{c}{Graph Accuracy} \\ \hline
     & \multicolumn{2}{c}{Random}                                 
     & \multicolumn{1}{c}{Mean MCD} 
     & \multicolumn{1}{c}{MCD1} 
     & \multicolumn{1}{c}{MCD2} 
     & MCD3   \\ \hline
     & \multicolumn{1}{c}{train}
     & \multicolumn{1}{c}{test} 
     & \multicolumn{1}{c}{test}    
     & \multicolumn{1}{c}{test} 
     & \multicolumn{1}{c}{test} 
     & \multicolumn{1}{c}{test} \\ \hline
      \multicolumn{1}{l|}{HPD $\spadesuit$} &  - & - & 67.3 ($\mp$4.1)	 & 72.0 ($\mp$7.5)	 & 66.1 ($\mp$6.4) & 	63.9 ($\mp$5.7) \\
        \multicolumn{1}{l|}{HPD w/o Hierarchical Mechanism $\spadesuit$} &  - & - & -  & 21.3	 & 6.4  & 	10.1  \\
      \multicolumn{1}{l|}{T5-small + IR $\diamondsuit$} &  - &  - & 47.9 & - & -& -\\
     \hline \hline
     \multicolumn{1}{l|}{LSTM + Attn $\heartsuit$} & - &  97.4 ($\mp$0.3) & 14.9 ($\mp$1.1)	& 28.9 ($\mp$1.8) &	5.0 ($\mp$0.8) &	10.8 ($\mp$0.6)                         \\
     \multicolumn{1}{l|}{Transformer $\heartsuit$}
     & -  &  98.5 ($\mp$0.2) & 17.9 ($\mp$0.9) & 34.9 ($\mp$1.1) & 	8.2 ($\mp$0.3) & 10.6 ($\mp$1.1) \\

     \multicolumn{1}{l|}{Universal Transformer $\heartsuit$}   &  - & 98.0 ($\mp$0.3) & 18.9 ($\mp$1.4)	& 37.4 ($\mp$2.2) &	8.1 ($\mp$1.6) &	11.3 ($\mp$0.3) \\
     \multicolumn{1}{l|}{Evol. Transformer $\clubsuit$}   & - &  - & 20.8 ($\mp$0.7) &	42.4 ($\mp$1.0) & 9.3 ($\mp$0.8) & 10.8 ($\mp$0.2) \\
     \multicolumn{1}{l|}{LSTM + Simplified SPARQL $\spadesuit$}
     & - & - & 26.1 & 42.2 & 14.5 & 21.5 \\
     \multicolumn{1}{l|}{Transformer + Simplified SPARQL $\spadesuit$} & - & -  &  31.4 &  53.0 &19.5 & 21.6 \\
     \multicolumn{1}{l|}{T5-small from scratch $\diamondsuit$}     & -  & -  &  20.8 & -& -& -\\
     \multicolumn{1}{l|}{T5-small from scratch + IR $\diamondsuit$} &  - &   - & 22.6 & -& -& -\\
\hline
\multicolumn{1}{l|}{Transformer$_{sh}$ weakly sup. LAGr, $K=1$}     &100 ($\mp$0.0) &  99.5 ($\mp$0.2) & \textbf{38.2} ($\mp$2.7) & \textbf{65.2} ($\mp$2.6)  & 26.4 ($\mp$3.2) & 23.0 ($\mp$2.0) 
\\
\multicolumn{1}{l|}{Transformer$_{sh}$ weakly sup. LAGr, $K=5, \sigma=10$}  & 100 ($\mp$0.0)  & 99.7 ($\mp$0.0) & \textbf{39.5} ($\mp$3.2) & 62.8 ($\mp$4.0) & \textbf{30.3} ($\mp$2.7) & \textbf{25.4} ($\mp$2.7)
  \\
\hline
\end{tabular}
\end{adjustbox}
\caption{Average graph accuracy and standard deviation of weakly-supervised LAGr on CFQ (\textbf{bottom}). \textbf{Middle:} results by several seq2seq baselines from prior work (\citet{keysers_measuring_2020} $\heartsuit$, \citet{furrer_compositional_2020} $\clubsuit$ ). \textbf{Top:} results not directly comparable to LAGr: Hierarchical Poset Decoding \cite{guo_hierarchical_2020} $\spadesuit$, and pretrained T5-small seq2seq model with intermediate representations (IR) \cite{herzig_unlocking_2021} $\diamondsuit$. Approaches other than LAGr report the average exact match accuracy with 95\% confidence intervals.
}
\label{cfq_results}
\end{table*}

\label{cfq_experiments}
\xhdr{Dataset} CFQ \citep{keysers_measuring_2020} is a benchmark for systematic generalization in semantic parsing that requires models to translate English sentences to SPARQL database queries.
We use CFQ's \textit{Maximum Compound Divergence} (MCD) splits, which were generated by making the distribution of compositional structures in the train and test sets as divergent as possible.

SPARQL queries contain two components: a \texttt{SELECT} and a \texttt{WHERE} clause.
The \texttt{SELECT} clause is either of the form \texttt{SELECT count(*)} for yes/no questions or \texttt{SELECT DISTINCT ?x0} for wh- questions (those starting with "which", "what", "who", etc.).
The \texttt{WHERE} clause can contain constrains of three kinds: filter constraints ensuring two variables or entities are distinct (e.g. \texttt{FILTER ?x0 != M0}), two-place predicates expressing a relation between two entities (e.g. \texttt{?x0 parent ?x1}), and one-place predicates expressing if an entity belongs to a category (e.g. \texttt{?x0 a ns:film.actor})

\xhdr{Graph Construction}
Before constructing the graphs, similarly to prior work \citep{furrer_compositional_2020, guo_hierarchical_2020}, we compress the SPARQL queries by merging some triples in the \texttt{WHERE} clauses. As an example, consider the question ``\textit{Were M2 and M3 directed by a screenwriter that executive produced M1?}'', where the original MR contains both \texttt{[M2 directed\_by ?x0, M3 directed\_by ?x0]} conjuncts. 
To make it easier to align SPARQL queries to the input question, we merge triples by concatenating their subjects and objects, e.g. yielding \texttt{[[M2, M3] directed\_by ?x0]} for the above example. 
With this compression, the  SPARQL queries can now contain an arbitrary number of entities in the triples. 
To convert the compressed SPARQL queries to graphs we first remove the \texttt{SELECT} clauses. To preserve the question type information, for wh- questions we replace the \texttt{?x0} variable in the \texttt{WHERE} clause with a special \texttt{select\_?x0} variable. 
As the example in Figure \ref{aligned_graph:1b} shows, we define the graph nodes by taking the entities (including variables, e.g. \texttt{?x0, M1}) and all predicates (\texttt{parent, sibling, actor}) from the triples.
For one-place predicates, we connect the entity nodes to the predicate node with an \texttt{agent} edge label. For triples with two-place predicates, we connect the predicate to the left-hand side and right-hand side entities with the \texttt{agent} and \texttt{theme} edge respectively. 
We add a \texttt{FILTER} edge between the variables or entities that participate in a filter constraint.
The resulting node and the edge vocabularies contain 84 and 4 labels respectively, each also including a \texttt{null} label.

\xhdr{Training Details}
Unlike COGS, we use L=2 graph layers in LAGr in order to accommodate for the larger MR graphs in CFQ.
This is because CFQ contains examples such as ``\textit{Who married M1's female German executive producer?}'' that contains 8 tokens, but induces the following 10 nodes:\texttt{?x1, executive\_produced, M1, gender, ns:m.02zsn, nationality, ns:m.0345h, select\_?x0, spouses, person}.

In all our CFQ experiments we use a shared Transformer encoder for both node and edge prediction. To assess performance, we use exact graph accuracy, which we define as the percentage of examples where the predicted and true graphs are isomorphic.
The predicted graphs contain enough information to exactly reconstruct the SPARQL query, hence
our exact graph accuracy can be compared to the exact match accuracy from the prior work.
For hyperparameter tuning, we follow \citet{keysers_measuring_2020} and use CFQ's in-distribution \textit{random} split to find the best model configuration. We do this by first fixing the number of candidate alignments at $K=1$ to search for the best hyperparameters. Once we find the best configuration, we tune $K$ and $\sigma$.
For the best found configuration of $K=5, \sigma=10$, as well as for the base configuration $K=1,\sigma=0$, we report the average graph accuracy and standard deviation for 8-11 runs of weakly-supervised LAGr on the MCD1, MCD2, MCD3 and the random split. 
Similarly to COGS, we use the PED initialization technique from \citet{csordas2021devil}, and discard runs where weakly-supervised LAGr does not reach at least 99.5\% graph accuracy on the training set
(around 12\% of all runs).
For further details on our CFQ experiments we refer the reader to Appendix \ref{cfq_tuning}. 

\xhdr{Results}
We compare LAGr to seq2seq semantic parsing results reported in prior work \citep{keysers_measuring_2020, furrer_compositional_2020}, as well as results obtained with compressed SPARQL queries \citep{guo_hierarchical_2020,herzig_unlocking_2021}.
As shown in Table \ref{cfq_results}, weakly-supervised LAGr outperforms all comparable baselines on all of CFQ's out-of-distribution MCD splits.
While both $K=1$ and $K=5$ with $\sigma=10$ yield impressive performance gains compared to the baselines, we obtain mixed results about the impact of a higher K and the use of noise. 
Specifically, the best result on MCD1 is achieved with $K=1$ in contrast to MCD2 and MCD3 where $K=5$ with $\sigma=10$ performs significantly better than when using $K=1$.

For reference, Table \ref{cfq_results} also includes the state-of-the-art Hierarchical Poset Decoding \citep[HPD,][]{guo_hierarchical_2020} method (see Section \ref{sec:related_work}), which arguably is not a fair baseline to LAGr because of its use of sketch prediction and lexicons. Notably, when these techniques are not used, LAGr performs much better than their base HPD algorithm.

\begin{table}[t]
\centering
{\small 
\begin{tabular}{cc|cc}
        &       & \multicolumn{2}{c}{Graph Accuracy}  \\ \hline
$K$ & $\sigma$ & train           & test          \\ \hline
1       & 0.0   & 99.79 ($\mp$0.4)    & 98.75 ($\mp$0.5)  \\ 
5       & 0.01  & 99.92 ($\mp$0.1)    & 99.01 ($\mp$0.2)  \\
        & 0.1   & 99.88 ($\mp$0.1)    & 99.10 ($\mp$0.3)  \\
        & 1.0   & 99.85 ($\mp$0.2)    & 99.10 ($\mp$0.3)  \\
        & 10.0  & \textbf{99.97} ($\mp$0.1)   & \textbf{99.69} ($\mp$0.1) \\
        & 15.0  & 83.78 ($\mp$1.6)   & 83.73 ($\mp$1.7) \\
        & 20.0  & 2.18 ($\mp$0.17)    & 2.28 ($\mp$0.19)  \\ 
10      & 0.01  & 99.77 ($\mp$0.3)    & 98.85 ($\mp$0.6)  \\
        & 0.1   & 99.92 ($\mp$0.1)    & 99.10 ($\mp$0.2)   \\
        & 1.0   & 99.70 ($\mp$0.3)    & 98.68 ($\mp$0.7)   \\
        & 10.0  & 99.96 ($\mp$0.1)    & 99.58 ($\mp$0.2)  \\
        & 15.0  & 99.77 ($\mp$0.4)   & 99.42 ($\mp$0.5) \\
        & 20.0  & 69.69 ($\mp$3.9)   & 68.91 ($\mp$4.0) \\ \hline
\end{tabular}
\caption{The effect of the number of alignment candidates $K$ and noise level $\sigma$ on the performance of weakly-supervised LAGr using CFQ's random split. We report the average graph accuracy and the standard deviation over 5 runs. We show the best configuration in bold.}
\label{cfq_ablations}}
\end{table}

To further zoom into the impact of the weakly-supervised LAGr's hyperparameters, we report results of preliminary experiments\footnote{These experiments were carried out using an earlier preliminary implementation. Results in Table \ref{cfq_ablations} are thus not directly comparable to those reported in Table \ref{cfq_results}.} in which we tuned the number of alignment candidates $K$ and the noise level $\sigma$. One can see that choosing the best alignment out of $K > 1$ candidates is indeed helpful, and that noise of high magnitude ($\sigma = 10$) brings the best improvement on the random split. 
These improvements also translate into systematic generalization gains for MCD2 and MCD3, as shown in Table \ref{cfq_results} where we see that $K=5$ achieves better performance than $K=1$.
The positive effect of a larger $K$ on these splits is in line with our expectation since 3.7 - 5.7\% of examples in each CFQ split have at least two predicates with identical node labels, which can make it hard to align the MR graph to the input by looking at node labels only.
Interestingly, in contrast to our intuition, when using ten candidate alignments, the random split test performance is slightly worse than when using five.
We show examples of the node labels that weakly-supervised LAGr predicts in the learned aligned CFQ graphs as well as the corresponding SPARQL queries in \Cref{cfq_predictions} (\Cref{errors}).

\section{Discussion \& Future Work}
\label{sec:discussion}
In this work we have shown that performing semantic parsing by labeling aligned graphs brings significant gains in systematic generalization. 
In our COGS and CFQ experiments, LAGr significantly improves upon  sequence-to-sequence baselines in both strongly and weakly-supervised settings.
Specifically, on COGS, LAGr outperforms our carefully-tuned seq2seq baselines and performs similarly to LSTMs that leverage lexicons. Lexicons can also be integrated into LAGr, although we do not expect this to improve LAGr's performance on COGS, as our best performing models already predict node labels perfectly. Lexicons also bring their own challenges of dealing with context-dependency and ambiguity, hence it is notable that LAGr matches the performance of a lexicon-equipped model while making less assumptions about the nature of the input-to-output mapping. On CFQ, LAGr outperforms all seq2seq baselines on all MCD splits. 

Based on our error analysis (see Appendix \ref{errors}), we believe that a modification of LAGr that conditions edge predictions on node labels could bring further improvements. Importantly, this modification would be compatible with our current alignment inference algorithm. Another obvious direction to improve LAGr's performance is by using a pretrained encoder.
Lastly, while the current alignment inference algorithm is effective, applying more advanced discrete optimization or amortized inference methods could be an interesting direction for future work. 

\subsubsection*{Acknowledgments}
We are thankful to Joelle Pineau, Siva Reddy and Christopher Manning for early discussions on this project. Furthermore, we also thank Nitarshan Rajkumar, Torsten Scholak and the rest of the Human-Machine Interaction Through Language group at ServiceNow for their invaluable feedback, reviews and contributions to this paper.
This research was supported in part by Canada CIFAR AI Chairs held by Prof. Pineau and Prof.Hamilton, as well as gift grants from Microsoft Research and Samsung AI. 

\bibliography{dima,dora}
\bibliographystyle{acl_natbib}

\appendix
\section{Appendix}

\subsection{COGS Hyperparameter Tuning}
\label{cogs_gen_dev}
COGS does not include an out-of-distribution development set, which makes it challenging to find the best model configuration.
To overcome this problem, we followed the same hyperparameter tuning procedure for our baselines and our strongly-supervised LAGr models as proposed by \citet{conklin_meta-learning_2021}. 
We sampled 1000 examples from the generalization set as a "Gen Dev" set which was used to pick the best hyperparameter configuration.
We tested  0.001, 0.004, 0.0001 and 0.0004 for learning rates, 64, 128 and 256 for batch sizes, and 0.1 versus 0.4 for dropout. We tested an embedding size of 256 versus 512. Furthermore, for the Transformer baselines and for LAGr with a Transformer encoder, we also tested 2 versus 4 layers, and 4 versus 8 attention heads. 
We trained all models for 70,000 steps, with no early stopping. 

Each configuration was evaluated on 5 seeds. Once the best configuration was found, we retrained all models on at least 10 seeds. 
The final number of seeds that were used to report our results in Table \ref{cogs_results} are the following: 20 seeds for each of the weakly-supervised LAGr experiments with and without retraining, 80 and 20 seeds for strongly-supervised LAGr with a separate and shared encoder, respectively, and finally, 20 seeds for our baseline Transformer experiments. We varied the number of seeds in order to obtain more accurate estimates for the mean performance measures.
The best configurations for COGS are shown in Table \ref{best_cogs_hyperparams}.

For weakly-supervised LAGr, we used the best configuration we found for strongly-supervised LAGr. We then investigated different values for $K$, the number of candidate alignments, with 1, 5 versus 10, and for the noise levels $\sigma$ of 0, 0.001, 0.01, 0.1, 1, 10, 15 and 20. In addition, we also implemented a random restart procedure to restart runs with a new random seed if they were not able to reach at least 98\% of training accuracy. We found that only when we used $K=10$ with $\sigma=1$, we were able to get around 50\% of the runs to converge. This was different from our CFQ experiments, where 97\% of runs converged to at least 98\% when appropriate noise levels were chosen (i.e., $\sigma < 15$).

As for our seq2seq baseline, in order to reproduce the same Transformer performance as reported by \citet{csordas2021devil}, we reused both their hyperparameters and their model implementation. Namely, we used a learning rate of 1e-4 with a linear scheduler and no warmup, a batch size of 128, an encoder dimension of 512 with dropout of 0.1. Lastly, we clipped gradients larger than 1.0. 

\subsection{CFQ Hyperparameter Tuning}
\label{cfq_tuning}
We performed hyperparameter tuning on CFQ's random split, and chose the best configuration based on the development exact graph accuracy. For LAGr with both shared and separate Transformer encoders, we tested learning rates of 0.0001, 0.0004, 0.0006, 0.0008 and 0.001, with a linear warmup of 0, 1000 versus 5000 steps, with dropout of 0.1 and 0.4,  batch sizes of 64, 128, 256 and 512, and 2 versus 4 Transformer layers and attention heads of 4 versus 8.
In contrast to COGS, we were able to drive the training loss to 0 without caching and appending previously learned alignments as the $K+1$st alignment candidates. For this reason, we did not use this caching technique.
Lastly, similarly to COGS, we filtered out runs that diverged in terms of their training graph accuracy. While for COGS weakly-supervised LAGr is more sensitive to varying $K$ and $\sigma$, in CFQ, we obtained 97\% convergence from all our runs in Table \ref{cfq_ablations}.
We report the best configuration used for CFQ in Table \ref{cfq_best_config}.

\subsection{Error analysis}
\label{errors}
Table \ref{tab:cogs_errors} shows some commonly encountered errors on COGS with strongly-supervised LAGr. In all examples, the model predicted the correct set of nodes. However, even when all nodes are correctly predicted, some may not show up in the final logical form, if it has no connecting edges to other nodes (see the "dog" node in example 4.). 

Figure \ref{cfq_predictions} shows the predicted nodes of aligned graphs and resulting queries produced by the best weakly-supervised LAGr model on CFQ. The top two rows show common errors where some edge labels do not get predicted, and where some nodes are missing due to the model not having predicted any connecting edges for the nodes, thus omitting the nodes from the final output graph. The bottom two rows show the inferred aligned graphs for examples that result in the correct output graph.

\begin{table*}
\centering
\tiny
\setlength\tabcolsep{2pt}
\begin{tabularx}{\textwidth}{ll}
\multicolumn{2}{l}{\textbf{Example 1: wrong edge label, between right nodes}} \\ \hline
\multicolumn{1}{l|}{\textbf{In}}   & A cockroach sent Sophia the sandwich beside the yacht .\\ \hline
\multicolumn{1}{l|}{\textbf{Out}}  & \begin{tabular}[c]{@{}l@{}}* sandwich ( x \_ 5 ) ; * yacht ( x \_ 8 ) ; cockroach ( x \_ 1 ) AND send . \textbf{theme} ( x \_ 2 , x \_ 1 ) AND send . recipient ( x \_ 2 , Sophia ) \\AND send . theme ( x \_ 2 , x \_ 5 ) AND sandwich . nmod . beside ( x \_ 5 , x \_ 8 )\end{tabular}                                               \\
\multicolumn{1}{l|}{\textbf{Pred}} & \begin{tabular}[c]{@{}l@{}}* sandwich ( x \_ 5 ) ; * yacht ( x \_ 8 ) ; cockroach ( x \_ 1 ) AND send . \textbf{agent} ( x \_ 2 , x \_ 1 ) AND send . recipient ( x \_ 2 , Sophia ) \\AND send . theme ( x \_ 2 , x \_ 5 ) AND sandwich . nmod . beside ( x \_ 5 , x \_ 8 )\end{tabular}                                               \\ \hline
 & \\
\multicolumn{2}{l}{\textbf{Example 2: Right edge label, but between wrong nodes}} \\ \hline
\multicolumn{1}{l|}{\textbf{In}}   & The girl beside the bed lended the manager the leaf .   \\ \hline
\multicolumn{1}{l|}{\textbf{Out}} & \begin{tabular}[c]{@{}l@{}}* girl ( x \_ 1 ) ; * bed ( x \_ 4 ) ; * manager ( x \_ 7 ) ; * leaf ( x \_ 9 ) ; \textbf{girl . nmod . beside ( x \_ 1 , x \_ 4 )} AND lend . agent ( x \_ 5 , x \_ 1 ) \\AND lend . recipient ( x \_ 5 , x \_ 7 ) AND lend . theme ( x \_ 5 , x \_ 9 )\end{tabular}                                       \\
\multicolumn{1}{l|}{\textbf{Pred}} & \begin{tabular}[c]{@{}l@{}}* girl ( x \_ 1 ) ; * bed ( x \_ 4 ) ; * manager ( x \_ 7 ) ; * leaf ( x \_ 9 ) ; lend . agent ( x \_ 5 , x \_ 1 ) \\AND lend . recipient ( x \_ 5 , x \_ 7 ) AND lend . theme ( x \_ 5 , x \_ 9 ) AND \textbf{leaf . nmod . beside ( x \_ 9 , x \_ 4 ) }\end{tabular}                                      \\ \hline
&\\
\multicolumn{2}{l}{\textbf{Example 3: Mistaking edge labels}}  \\ \hline
\multicolumn{1}{l|}{\textbf{In}}   & The dog noticed that a hippo juggled . \\ \hline
\multicolumn{1}{l|}{\textbf{Out}}  & * dog ( x \_ 1 ) ; notice . agent ( x \_ 2 , x \_ 1 ) AND notice . ccomp ( x \_ 2 , x \_ 6 ) AND hippo ( x \_ 5 ) AND juggle . \textbf{agent} ( x \_ 6 , x \_ 5 ) \\
\multicolumn{1}{l|}{\textbf{Pred}} & * dog ( x \_ 1 ) ; notice . agent ( x \_ 2 , x \_ 1 ) AND notice . ccomp ( x \_ 2 , x \_ 6 ) AND hippo ( x \_ 5 ) AND juggle . \textbf{theme} ( x \_ 6 , x \_ 5 ) \\ \hline & \\
\multicolumn{2}{l}{\textbf{Example 4: Correct nodes, but incorrect edges predicted}} \\ \hline
\multicolumn{1}{l|}{\textbf{In}}   & A dog beside a chair said that a melon on the bed was liked . \\ \hline
\multicolumn{1}{l|}{\textbf{Out}}  & \begin{tabular}[c]{@{}l@{}}* bed ( x \_ 11 ) ; dog ( x \_ 1 ) AND \textbf{dog . nmod . beside ( x \_ 1 , x \_ 4 )} AND chair ( x \_ 4 ) AND say . agent ( x \_ 5 , x \_ 1 ) \\AND say . ccomp ( x \_ 5 , x \_ 13 ) AND  melon ( x \_ 8 ) AND \textbf{melon} . nmod . on ( x \_ 8 , x \_ 11 ) AND like . theme ( x \_ 13 , x \_ 8 )\end{tabular} \\
\multicolumn{1}{l|}{\textbf{Pred}} & \begin{tabular}[c]{@{}l@{}}* bed ( x \_ 11 ) ; chair ( x \_ 4 ) AND say . agent ( x \_ 5 , x \_ 4 ) AND melon ( x \_ 8 ) AND \textbf{bed} . nmod . in ( x \_ 11 , x \_ 13 ) \\AND like . theme ( x \_ 13 , x \_ 8 )\end{tabular}
\end{tabularx}
\caption{Incorrectly predicted logical forms for COGS with strongly-supervised LAGr. Errors are highlighted in bold.}
\label{tab:cogs_errors}
\end{table*}

\subsection{Further COGS examples}
\label{cogs_examples}
Table \ref{cogs_examples_table} shows further examples from COGS's generalization set with various cases for challenging models' ability to test systematic generalization.

\begin{figure*}
\begin{subtable}[H]{\textwidth}
\tiny
\setlength\tabcolsep{2pt}
    \begin{tabularx}{0.9\textwidth}{l|l|l|l|l|l|l|l|l|l|l|l|l|l|l|l|l|l|l|l}
    \multicolumn{20}{l}{Example 1: Wrong edge predictions}\\ \hline
    \textbf{Layer 2}& ?x0               & M3                & influenced        &                  &              & director       &                       & spouse            & M2        & ?x2              & cinematographer       &                              &                 & M4        &                  &           &                  & ?x1       & actor \\
    \hline
    \textbf{Layer 1} & & & & & & & & & & & & & & & & & & &\\ \hline
    \textbf{Input} & Did & M3 & influence & a & film & director & , & marry & M2 & 's & cinematographer & , & influence & M4 & , & and & influence & a & actor\\
    \hline
    \multicolumn{1}{l|}{\textbf{Target}}              & \multicolumn{19}{l}{?x1 actor   .   ?x0 director  .   ?x2 cinematographer M2   .   FILTER M3 != ?x2   .   M3 influenced {[}?x0 ?x1 M4{]}   .   M3 spouse ?x2}\\
    \multicolumn{1}{l|}{\textbf{Predicted}}           & \multicolumn{19}{l}{?x0 actor   .   ?x0 director   .   ?x1 director   .   ?x2 cinematographer M2   .   FILTER M3 != ?x2   .   M3 influenced {[}?x0 ?x1 M4{]}   .   M3 spouse ?x2}\\
    \hline
    \end{tabularx}
\end{subtable}

\vspace*{5pt}

\begin{subtable}[H]{\textwidth}
\tiny
\setlength\tabcolsep{1.5pt}
    \begin{tabularx}{\textwidth}{l|l|l|l|l|l|l|l|l|l|l|l|l|l|l|l|l|l}
    \multicolumn{18}{l}{Example 2: Missing node}\\
    \hline
    \textbf{Layer 2} & select\_?x0 & ns:m.0f8l9c & & editor & & M1 & influenced\_ by & & ?x1 & employer & ?x2 & organizations\_founded & & & & & M2\\
    \hline
    \textbf{Layer 1} & & nationality & &&&&&&&&&&&&&&\\
    \hline
    \textbf{Input} & What & French & film & editor & that & M1 & influenced & influenced & a & company & s & founder & and & was & influenced & by & M2\\
    \hline
    \multicolumn{1}{l|}{\textbf{Target}} & \multicolumn{17}{l}{?x1 actor   .   ?x0 director  .   ?x2 cinematographer M2   .   FILTER M3 != ?x2   .   M3 influenced {[}?x0 ?x1 M4{]}   .   M3 spouse ?x2}\\
    \multicolumn{1}{l|}{\textbf{Predicted}} & \multicolumn{17}{l}{?x0 actor   .   ?x0 director   .   ?x1 director   .   ?x2 cinematographer M2   .   FILTER M3 != ?x2   .   M3 influenced {[}?x0 ?x1 M4{]}   .   M3 spouse ?x2}\\ 
    \hline
    \end{tabularx}
\end{subtable}

\vspace*{5pt}

\begin{subtable}[c]{0.55\textwidth}
\tiny
\setlength\tabcolsep{2pt}
    \begin{tabularx}{\textwidth}{l|l|l|l|l|l|l|l}
    \multicolumn{8}{l}{Example 3: Correct prediction}\\ \hline
    \textbf{Layer 2} & select\_?x0       & ns:m.05zppz       & ns:m.059j2 & & editor & director & M3\\
    \hline
    \textbf{Layer 1} & & & gender & nationality & & &\\
    \hline
    \textbf{Input} & Which & male & Dutch & film & editor & directed & M3\\
    \hline
    \multicolumn{1}{l|}{\textbf{Predicted}} & \multicolumn{7}{X}{select\_?x0 director M3   .   select\_?x0 editor   .   select\_?x0 gender ns:m.05zppz   .   select\_?x0 nationality ns:m.059j2}\\
    \hline
    \end{tabularx}
\end{subtable}

\vspace*{5pt}

\begin{subtable}[c]{0.75\textwidth}
\tiny
\setlength\tabcolsep{2pt}
    \begin{tabularx}{\textwidth}{l|l|l|l|l|l|l|l|l|l|l|l|l}
    \multicolumn{13}{l}{Example 4: Correct prediction}\\ \hline
    \textbf{Layer 2 }& select\_?x0 && ns:m.06mkj & actor && influenced & M2 && ?x1 & actor &\\
    \hline
    \textbf{Layer 1 } & nationality & person &&&&&&&\\
    \hline
    \textbf{Input} & Who & was & a & Spanish & actor & that & influenced & M2 & and & influenced & a & actor\\
    \hline
    \multicolumn{1}{l|}{\textbf{Predicted}}           & \multicolumn{12}{X}{?x1 actor   .   select\_?x0 actor    .   select\_?x0 influenced ?x1    .   select\_?x0 influenced M2    .   select\_?x0 person    .   select\_?x0 nationality ns:m.06mkj}\\ 
    \hline
    \end{tabularx}
\end{subtable}

\caption{Predicted nodes of aligned graphs and resulting queries produced by the best weakly-supervised LAGr with $k=5$, $\sigma=10$ on the development set of CFQ. Top two rows show common errors with missing edge labels and missing nodes, and bottom rows show the inferred alignments for correct examples.}
\label{cfq_predictions}
\end{figure*}

\begin{table*}
\centering
\begin{tabular}{lll}
\hline
Case                           & Training                                                                               & Generalization                                                                                      \\ \hline
Subject $\rightarrow$ Object   & A \textbf{hedgehog} ate the cake.                                                               & The baby liked the \textbf{hedgehog}.                                                                        \\
Object $\rightarrow$ Subject   & Henry liked a \textbf{cockroach}.                                                               & The \textbf{cockroach} ate the bat.                                                                          \\
Primitive $\rightarrow$ Object & \textbf{Paula}                                                                                  & The child helped \textbf{Paula}.                                                                             \\
Depth generalization           & \begin{tabular}[c]{@{}l@{}}Ava saw the ball \textbf{in the bottle}\\ \textbf{on the table.}\end{tabular} & \begin{tabular}[c]{@{}l@{}}Ava saw the ball \textbf{in the bottle}\\ \textbf{on the table on the floor.}\end{tabular} \\
Active $\rightarrow$ Passive   & Emma \textbf{blessed} William.                                                                  & A child was \textbf{blessed}.                                                                                \\ \hline
\end{tabular}
\caption{Example from \citet{kim_cogs_2020} that show various linguistic phenomena from the COGS generalization set.}
\label{cogs_examples_table}
\end{table*}

\begin{table*}
\centering
\begin{adjustbox}{width=\textwidth}
\begin{tabular}{lcccccc}
& \multicolumn{2}{c}{\textbf{Reproduced baselines}}. & \multicolumn{4}{c}{\textbf{Strongly-supervised LAGr with different encoders}} \\ \hline
\multicolumn{1}{c}{}. & LSTM    & \multicolumn{1}{c|}{Transformer} & $\text{LSTM}_{sh}$ & $\text{LSTM}_{sep}$  & $\text{Transformer}_{sh}$  & $\text{Transformer}_{sep}$  \\ \hline
\multicolumn{1}{l|}{batch\_size} & 256 & \multicolumn{1}{c|}{128} & 128 & 64 & 128 & 128\\
\multicolumn{1}{l|}{learning\_rate} & 0.004& \multicolumn{1}{c|}{0.0001} & 0.0001 & 0.0004 & 0.0001 & 0.0001\\
\multicolumn{1}{l|}{scheduler}                                                        & \begin{tabular}[c]{@{}c@{}}linear with \\ warmup of 1000 steps\end{tabular} & 
    \multicolumn{1}{c|}{\begin{tabular}[c]{@{}c@{}}linear with \\ no warmup \end{tabular}}
& \begin{tabular}[c]{@{}c@{}}linear with \\ warmup of 1000 steps \end{tabular}
& \begin{tabular}[c]{@{}c@{}}linear with \\ warmup of 1000 steps \end{tabular} 
& \begin{tabular}[c]{@{}c@{}}linear with \\ no warmup \end{tabular} 
& \begin{tabular}[c]{@{}c@{}}linear with \\ no warmup \end{tabular} \\
\multicolumn{1}{l|}{layers} & 2 & \multicolumn{1}{c|}{4} & 2 & 2 & 4 & 4 \\
\multicolumn{1}{l|}{enc\_dim}  & 256 & \multicolumn{1}{c|}{256} & 256 & 256 & 512 & 512                                                                         \\
\multicolumn{1}{l|}{train\_steps}                                                            & 50000 & \multicolumn{1}{c|}{50000} 
& 70000 & 70000 & 70000 & 70000                                                                       \\
\multicolumn{1}{l|}{dropout} & 0.4 & \multicolumn{1}{c|}{0.1} & 0.1 & 0.4 & 0.4 & 0.4 \\
\multicolumn{1}{l|}{attention heads}  & - & \multicolumn{1}{c|}{8} & -  & - & 4 & 4 \\                               
\hline
\end{tabular}
\end{adjustbox}
\caption{Best hyperparameters for our COGS baseline and strongly-supervised LAGr experiments}
\label{best_cogs_hyperparams}
\end{table*}

\begin{table*}
\centering
\begin{tabular}{lcc}
\multicolumn{1}{c}{}                                                                           & \multicolumn{2}{c}{CFQ} \\ \hline
& \multicolumn{2}{c}{\textbf{Weakly-supervised LAGr}}                                         \\ \hline
\multicolumn{1}{c}{}                                                                           & Transformer$_{sh}$                                                         \\ \hline
\multicolumn{1}{l|}{batch\_size}                                                               & 256                                                                         \\
\multicolumn{1}{l|}{learning\_rate}                                                            & 0.0004                                                                      \\
\multicolumn{1}{l|}{scheduler}                                                                 & \begin{tabular}[c]{@{}c@{}}linear with warmup \\ of 1000 steps\end{tabular} \\
\multicolumn{1}{l|}{layers}                                                                     & 4                                                                           \\
\multicolumn{1}{l|}{enc\_dim}                                                                   & 256                                                                         \\
\multicolumn{1}{l|}{train\_steps}                                                               & 750000                                                                      \\
\multicolumn{1}{l|}{dropout}                                                                    & 0.1                                                                         \\
\multicolumn{1}{l|}{\begin{tabular}[c]{@{}l@{}}attention\\ heads\end{tabular}}                  & 8
\\ \hline
\end{tabular}
\caption{Best configuration for CFQ weakly-supervised LAGr.}
\label{cfq_best_config}
\end{table*}

\end{document}

%% file: math_commands.tex
\usepackage{amsmath,amsfonts,bm}

\def\eqref#1{equation~\ref{#1}}

\def\1{\bm{1}}

\DeclareMathAlphabet{\mathsfit}{\encodingdefault}{\sfdefault}{m}{sl}
\SetMathAlphabet{\mathsfit}{bold}{\encodingdefault}{\sfdefault}{bx}{n}

\newcommand{\softmax}{\mathrm{softmax}}

\DeclareMathOperator*{\argmax}{arg\,max}